\documentclass[journal]{IEEEtran}
\usepackage{cite}
\usepackage{amsmath,amssymb,amsfonts}
\usepackage{algorithmic}
\usepackage{graphicx}
\usepackage{algorithm}
\usepackage{hyperref}
\usepackage{tikz}
\usetikzlibrary{positioning}
\usepackage{xcolor}
\hypersetup{hidelinks=true}
\usepackage{textcomp}
\usepackage{booktabs,siunitx,subcaption}
\usepackage{multirow}
\usepackage{adjustbox}
\usetikzlibrary{positioning,arrows.meta,calc}
\usepackage{array}

\newcommand{\textcc}[1]{\textsc{#1}}

\def\BibTeX{{\rm B\kern-.05em{\sc i\kern-.025em b}\kern-.08em
    T\kern-.1667em\lower.7ex\hbox{E}\kern-.125emX}}
\begin{document}

\title{MMCTOP: A Multimodal Textualization and Mixture-of-Experts Framework for Clinical Trial Outcome Prediction}

\author{Carolina Aparício, Qi Shi, Bo Wen, Tesfaye Yadete, and Qiwei Han%
% 1. Funding
\thanks{%Manuscript received Month DD, YYYY; revised Month DD, YYYY. 
This work was funded by Fundação para a Ciência e a Tecnologia (UIDB/00124/2020, UIDP/00124/2020 and Social Sciences DataLab - PINFRA/22209/2016), POR Lisboa and POR Norte (Social Sciences DataLab, PINFRA/22209/2016). Bo Wen and Tesfaye Yadete acknowledge support from the Cleveland Clinic - IBM Discovery Accelerator.}%
% 2. Carolina & Qi
\thanks{C. Aparício and Q. Shi are with Nova School of Business and Economics, Carcavelos, Portugal (e-mail: 61582@novasbe.pt; qi.shi@novasbe.pt).}%
% 3. Bo Wen (Current + Past)
\thanks{B. Wen is with Hogarthian Technologies, New York, NY, USA (e-mail: bwen@hogarthian.com). He was with IBM Research, Yorktown Heights, NY, USA, when this work was performed.}%
% 4. Tesfaye Yadete (Current + Past)
\thanks{T. Yadete is with the School of Medicine, Oregon Health \& Science University, Portland, OR, USA (e-mail: yadete@ohsu.edu). He was with the Cleveland Clinic, Cleveland, OH, USA, when this work was performed.}%
% 5. Qiwei Han (Corresponding)
\thanks{Q. Han is with Nova School of Business and Economics, Carcavelos, Portugal (corresponding author to provide e-mail: qiwei.han@novasbe.pt).}%
}

\maketitle

\begin{abstract}
Addressing the challenge of multimodal data fusion in high-dimensional biomedical informatics, we propose MMCTOP, a \textbf{M}ulti\textbf{M}odal \textbf{C}linical-\textbf{T}rial \textbf{O}utcome \textbf{P}rediction framework that integrates heterogeneous biomedical signals spanning (i) molecular structure representations, (ii) protocol metadata and long-form eligibility narratives, and (iii) disease ontologies. MMCTOP couples schema-guided textualization and input-fidelity validation with modality-aware representation learning, in which domain-specific encoders generate aligned embeddings that are fused by a transformer backbone augmented with a drug--disease--conditioned sparse Mixture-of-Experts (SMoE). This design explicitly supports specialization across therapeutic and design subspaces while maintaining scalable computation through top-\emph{k} routing. MMCTOP achieves consistent improvements in precision, F1, and AUC over unimodal and multimodal baselines on benchmark datasets, and ablations show that schema-guided textualization and selective expert routing contribute materially to performance and stability. We additionally apply temperature scaling to obtain calibrated probabilities, ensuring reliable risk estimation for downstream decision support. Overall, MMCTOP advances multimodal trial modeling by combining controlled narrative normalization, context-conditioned expert fusion, and operational safeguards aimed at auditability and reproducibility in biomedical informatics.
\end{abstract}


\section{Introduction} \label{sec:Introduction}

Clinical trials are the cornerstone of biomedical innovation, providing the mechanism through which new drugs and treatments are rigorously vetted for safety and efficacy before reaching patients. Yet the process remains plagued by inefficiencies, escalating costs, and low success rates; each failed study is a double loss, consuming scarce resources and delaying access to effective therapies \cite{sertkaya2016costs,sertkaya2024costs}. The stakes are enormous: the global pharmaceutical industry, valued at \$390~billion in 2000, has grown to exceed \$1.5~trillion by 2024 \cite{statista2023pharma}, intensifying pressure to streamline clinical development. Regulators, including the U.S.\ Food and Drug Administration (FDA), continue to emphasize improving the predictability and efficiency of clinical research to accelerate patient access \cite{woodcock2020efficiency,fda2025}.

Typically, clinical trial development proceeds through a structured, multi-phase pathway. Phase~I trials evaluate pharmacokinetics, tolerability, and initial safety in small cohorts; Phase~II trials assess preliminary efficacy alongside continued safety monitoring; and Phase~III trials conduct large-scale, controlled evaluations against standard-of-care or placebo with heightened statistical rigor \cite{fda2025}. Although early-phase studies may show promising signals, a substantial fraction of programs fail in Phases~II or III, where both financial and temporal stakes are highest. Late-stage failures are particularly costly: out-of-pocket expenditures for a single Phase~III trial commonly range from \$11.5 to \$52.9~million, with additional downstream costs arising from extended timelines and delayed market entry \cite{sertkaya2016costs,sertkaya2024costs}. The ability to anticipate trial outcomes, both within individual phases and across the full development trajectory, is therefore critical for portfolio de-risking, resource reallocation, and evidence-based decision-making in biomedical R\&D \cite{pmlr-v106-qi19a}.

In particular, clinical trial planning and evaluation increasingly depend on \emph{multimodal} biomedical evidence. Structured representations, such as disease ontologies and diagnostic codes, provide standardized descriptions of indications and comorbidities; unstructured protocol narratives and eligibility criteria specify populations, endpoints, and operational constraints; and molecular representations encode mechanism-of-action, chemical feasibility, and pharmacological properties \cite{rajkomar2019mlmedicine,miotto2016deep}. The integration of these heterogeneous data sources not only underpins recent advances in biomedical and health informatics, but also introduces a set of recurring challenges: (i) \emph{high-dimensional and heterogeneous inputs at scale}, (ii) \emph{representation mismatch} across graphs, hierarchies, tabular attributes, and long-form text, and (iii) \emph{privacy and security risks} associated with fusing sensitive clinical descriptors and derived representations. Addressing these challenges requires end-to-end pipelines that are not only accurate, but also can reconcile heterogeneous modality encoders and representations within a coherent fusion framework rather than assuming a single universal input form.

Early predictive models relied on hand-crafted features derived from narrow modalities such as pharmacokinetics or toxicity \cite{pmlr-v106-qi19a,gayvert2016datadriven,tropsha2010qsar}, and later incorporated trial-design elements (e.g., cohort size, comparators, endpoints) and broader biomedical context \cite{Lo2019Machine,wu2018moleculenet,tasneem2012aact}. While these models demonstrated feasibility, they often overfit to specific diseases or trial phases and struggled to generalize across indications, data modalities, and clinical settings \cite{zhou2022gnn,kappen2018evaluating}. These limitations stem in part from reliance on rigid fusion pipelines and tightly coupled encoders, which complicate extensibility when integrating new modalities or adapting to evolving biomedical knowledge.

Advances in deep learning have broadened the methodological toolkit. Graph neural networks (GNNs) model drug--disease--target interactions through heterogeneous biomedical graphs \cite{himmelstein2017systematic}; transfer learning adapts large-scale language models pretrained on biomedical corpora (e.g., BioBERT) for understanding protocol documents and clinical literature \cite{lee2020biobert}; and ensemble learning aggregates diverse predictors to improve robustness across treatment areas \cite{sahoo2021survey}. Despite this progress, multimodal integration remains an open challenge. Molecules are most naturally represented as graphs or SMILES sequences, diseases as ontological hierarchies, and trial protocols as long unstructured narratives; reconciling these representations within a unified predictive system introduces mismatches across structure, semantics, and granularity that complicate fusion, hinder onboarding of new modalities, and weaken cross-phase generalization \cite{miotto2016deep,shickel2018deepehr}.

Large Language Models (LLMs) offer a complementary paradigm for addressing some of these challenges by supporting \emph{textualization} and normalization of heterogeneous biomedical inputs into structured natural-language artifacts \cite{lin2025bridging}. In practice, however, robust clinical trial modeling continues to benefit from modality-appropriate encoders: molecular signals are most effectively captured by chemical language or structure models, while eligibility criteria and protocol narratives are better represented using biomedical language models. Accordingly, LLM-based textualization is best understood as an upstream \emph{engineering and governance layer} that standardizes inputs, enforces schema conformance, and enables cross-field consistency checks, rather than as a replacement for domain-specific representation learning. Recent frameworks such as HINT and LIFTED illustrate the promise of integrating multimodal biomedical signals using hierarchical interaction modeling and expert routing \cite{fu2022hint,zheng2025lifted}. Nevertheless, practical gaps remain in constructing \emph{auditable, schema-conformant, and clinically grounded} pipelines that explicitly separate input normalization from representation learning, while supporting scalable and context-aware fusion across heterogeneous modalities.

Importantly, clinical trial success is jointly constrained by \emph{biological plausibility} and \emph{operational design}. Molecular properties determine whether a compound can realistically modulate its intended target and disease pathway, while protocol structure governs patient selection, endpoints, and execution feasibility. MMCTOP explicitly bridges this divide by integrating bioinformatics-level molecular representation learning  with clinical informatics signals from trial protocols and eligibility narratives, enabling outcome prediction that reflects both pharmacological feasibility and trial design realism.

To address these challenges, we propose \textbf{MMCTOP}, a \textbf{M}ulti\textbf{M}odal \textbf{C}linical-\textbf{T}rial \textbf{O}utcome \textbf{P}rediction framework that standardizes heterogeneous trial signals via schema-guided textualization and grounding, encodes each modality using domain-specific transformers, and models cross-modality interactions using a transformer backbone augmented with a sparse Mixture-of-Experts (SMoE) mechanism. This design enables conditional specialization across drug--disease and protocol contexts while maintaining modularity through aligned embeddings and cached representations.

More specifically, MMCTOP introduces three core design components: 
(i) a \emph{schema-guided textualization and normalization} process that converts molecular structures, trial protocols, and disease ontologies into standardized narratives and slot-ordered fields with explicit validation constraints; 
(ii) automated \emph{input fidelity controls} based on schema conformance and cross-field validation (e.g., phase--enrollment coherence and ontology parent--child consistency), implemented as engineering safeguards for auditability; 
and (iii) a \emph{drug--disease--conditioned sparse gating} mechanism that routes aligned multimodal embeddings through specialized experts to achieve scalable and clinically meaningful specialization.

The paper presents the following contributions:
\vspace{-0.25em}
\begin{itemize}
  \item \textbf{Multimodal unification for fusion at scale:} a schema-guided textualization and validation procedure that standardizes trial attributes, protocol, ontology codes, and molecular descriptors into auditable artifacts with deterministic decoding and quality-control checks.
   \item \textbf{Cross-modality learning for trial outcomes:} a multimodal architecture that combines domain-specific encoders with transformer-based fusion and sparse expert routing to capture drug--disease and protocol interactions.
\end{itemize}

Although demonstrated on clinical trial outcome prediction, MMCTOP generalizes to other biomedical fusion tasks that require integrating structured and textual modalities, such as therapeutic response modeling, pharmacogenomic prediction, and patient stratification, without redesigning the core fusion architecture.

\section{Background and Related Work} \label{sec:Background}

\subsection{Clinical Trial Outcome Prediction}
Predictive modeling for clinical trial outcomes has long been recognized as a core challenge in biomedical informatics. Early efforts relied on structured drug- or trial-level features with relatively simple classifiers. Quantitative structure–activity relationship (QSAR) lines of work posited that molecular structure largely determines downstream efficacy/toxicity, correlating chemical descriptors with biological activity \cite{tropsha2010qsar}. Pharmacokinetic (PK) and pharmacodynamic proxies were also leveraged: subject-level or aggregated PK features (e.g., compound exposure) were used to extrapolate Phase~III effect sizes from Phase~II data \cite{pillai2024machine,vamathevan2019applications}. In parallel, toxicity-related properties and historical safety signals were associated with attrition risk and failures of clinical trials \cite{gayvert2016datadriven}.

Beyond drug-centric signals, researchers incorporated \emph{trial-design} features, such as cohort size and eligibility windows, randomization/blinding, comparator type (placebo vs.\ standard-of-care), arm count, and primary/secondary endpoints to capture procedural context that shapes outcomes \cite{Lo2019Machine}. Benchmark resources such as Aggregated Analysis of Clinical Trials (AACT) \cite{tasneem2012aact,ctti_aact} and molecular benchmarks like MoleculeNet \cite{wu2018moleculenet} enabled broader empirical studies across indications. These feature sets improved discrimination within specific phases or disease areas but often struggled to generalize, especially when moving from Phase~II to Phase~III where shifts in sample size, baseline risk, event rates, and endpoint definitions can alter measured effect sizes \cite{rajkomar2019mlmedicine,miotto2016deep}. This motivates frameworks that reason jointly over diverse biomedical signals and generalize across phase and indication shifts \cite{zhou2022gnn,kappen2018evaluating}.

\subsection{Advances in Multimodal Learning in Biomedical Informatics}
The prevalence of public resources has enabled learning from complementary modalities at scale: ClinicalTrials.gov/AACT contribute protocol metadata and rich narratives; DrugBank/PubChem/ChEMBL contribute molecular knowledge (identifiers, SMILES, substructures, mechanisms-of-action) \cite{wishart2018drugbank,kim2021pubchem,chembl,pubchem}; and biomedical ontologies such as MeSH/ICD/UMLS encode hierarchical disease semantics and terminology mappings \cite{bodenreider2004unified,who_icd10}. Aggregations such as the TOP and CTOD datasets curate historical outcomes for benchmarking \cite{gao2024automatically}. 

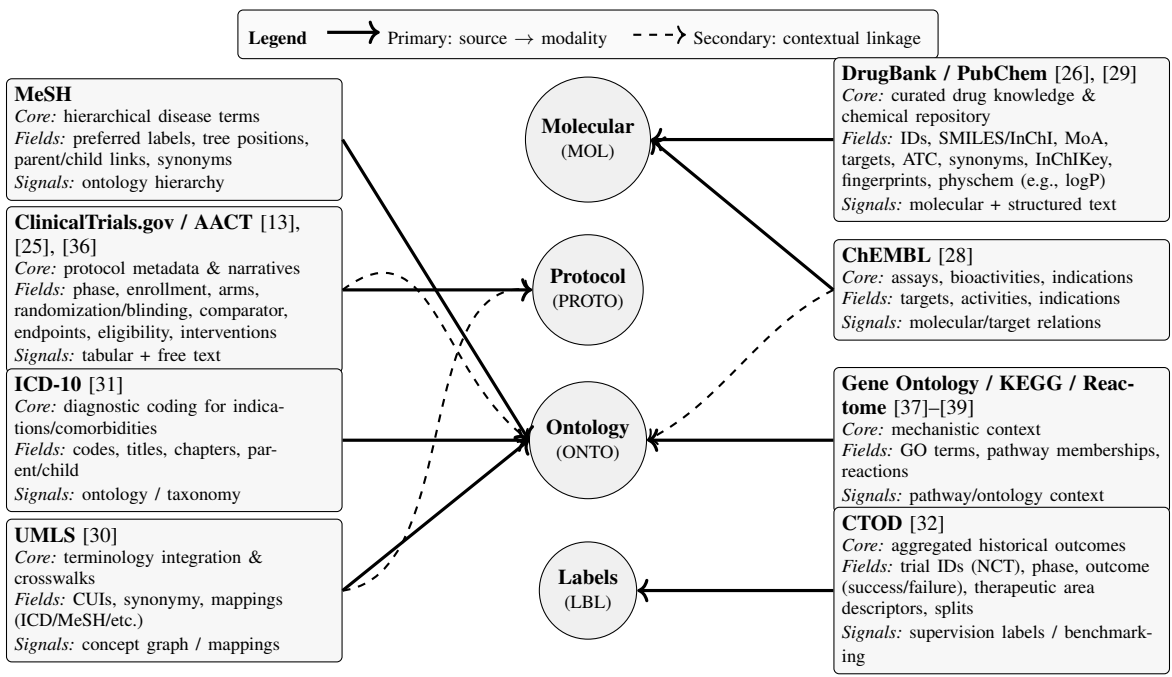
\begin{figure*}[!t]
\centering
\begin{adjustbox}{max width=\textwidth,center}
\begin{tikzpicture}[
  every node/.style={font=\footnotesize},
  resource/.style={draw, rounded corners=2pt, align=left, text width=4.25cm, minimum height=1.05cm, fill=gray!6, inner sep=3pt},
  modality/.style={draw, circle, align=center, minimum size=11mm, fill=gray!12},
  primary/.style={->, very thick},
  secondary/.style={->, dashed, thick},
  legendbox/.style={draw, rounded corners=2pt, inner sep=4pt, align=left, fill=gray!4}
]
\coordinate (C) at (0,0);
\node[modality] (mol)   at ($(C)+(0,  2.6)$) {\textbf{Molecular}\\ \scriptsize (MOL)};
\node[modality] (proto) at ($(C)+(0,  0.6)$) {\textbf{Protocol}\\ \scriptsize (PROTO)};
\node[modality] (onto)  at ($(C)+(0, -1.4)$) {\textbf{Ontology}\\ \scriptsize (ONTO)};
\node[modality] (lbls)  at ($(C)+(0, -3.4)$) {\textbf{Labels}\\ \scriptsize (LBL)};
\node[legendbox] (legend) at ($(mol)+(0,14mm)$) {%
\scriptsize \textbf{Legend}\quad
\tikz{\draw[primary] (0,0)--(7mm,0);} \scriptsize Primary: source $\to$ modality \quad
\tikz{\draw[secondary] (0,0)--(7mm,0);} \scriptsize Secondary: contextual linkage
};
\def\LX{-55mm}
\def\RX{ 55mm}
\node[resource] (mesh)  at ($(mol)+(\LX,0)$) {\textbf{MeSH}\\
\scriptsize \textit{Core:} hierarchical disease terms\\
\scriptsize \textit{Fields:} preferred labels, tree positions, parent/child links, synonyms\\
\scriptsize \textit{Signals:} ontology hierarchy
};
\node[resource] (ctgov) at ($(proto)+(\LX,0)$) {\textbf{ClinicalTrials.gov / AACT} \cite{clinicaltrials,ctti_aact,tasneem2012aact}\\
\scriptsize \textit{Core:} protocol metadata \& narratives\\
\scriptsize \textit{Fields:} phase, enrollment, arms, randomization/blinding, comparator, endpoints, eligibility, interventions\\
\scriptsize \textit{Signals:} tabular + free text
};
\node[resource] (icd)   at ($(onto)+(\LX,0)$) {\textbf{ICD-10} \cite{who_icd10}\\
\scriptsize \textit{Core:} diagnostic coding for indications/comorbidities\\
\scriptsize \textit{Fields:} codes, titles, chapters, parent/child\\
\scriptsize \textit{Signals:} ontology / taxonomy
};
\node[resource] (umls)  at ($(lbls)+(\LX,0)$) {\textbf{UMLS} \cite{bodenreider2004unified}\\
\scriptsize \textit{Core:} terminology integration \& crosswalks\\
\scriptsize \textit{Fields:} CUIs, synonymy, mappings (ICD/MeSH/etc.)\\
\scriptsize \textit{Signals:} concept graph / mappings
};
\node[resource] (drugchem) at ($(mol)+(\RX,0)$) {\textbf{DrugBank / PubChem} \cite{wishart2018drugbank,pubchem}\\
\scriptsize \textit{Core:} curated drug knowledge \& chemical repository\\
\scriptsize \textit{Fields:} IDs, SMILES/InChI, MoA, targets, ATC, synonyms, InChIKey, fingerprints, physchem (e.g., logP)\\
\scriptsize \textit{Signals:} molecular + structured text
};
\node[resource] (chembl)   at ($(proto)+(\RX,0)$) {\textbf{ChEMBL} \cite{chembl}\\
\scriptsize \textit{Core:} assays, bioactivities, indications\\
\scriptsize \textit{Fields:} targets, activities, indications\\
\scriptsize \textit{Signals:} molecular/target relations
};
\node[resource] (pathways) at ($(onto)+(\RX,0)$) {\textbf{Gene Ontology / KEGG / Reactome} \cite{ashburner2000gene,kanehisa2017kegg,jassal2020reactome}\\
\scriptsize \textit{Core:} mechanistic context\\
\scriptsize \textit{Fields:} GO terms, pathway memberships, reactions\\
\scriptsize \textit{Signals:} pathway/ontology context
};
\node[resource] (ctod)     at ($(lbls)+(\RX,0)$) {\textbf{CTOD} \cite{gao2024automatically}\\
\scriptsize \textit{Core:} aggregated historical outcomes\\
\scriptsize \textit{Fields:} trial IDs (NCT), phase, outcome (success/failure), therapeutic area descriptors, splits\\
\scriptsize \textit{Signals:} supervision labels / benchmarking
};
\draw[primary] (mesh.east)     -- (onto.west);
\draw[primary] (ctgov.east)    -- (proto.west);
\draw[primary] (icd.east)      -- (onto.west);
\draw[primary] (umls.east)     -- (onto.west);
\draw[primary] (drugchem.west) -- (mol.east);
\draw[primary] (chembl.west)   -- (mol.east);
\draw[primary] (pathways.west) -- (onto.east);
\draw[primary] (ctod.west)     -- (lbls.east);
\draw[secondary] (ctgov.east) .. controls +(1.2,0.9)  and +(-1.2,0.9)  .. (onto.west);
\draw[secondary] (umls.east)  .. controls +(1.6,0.2)  and +(-1.6,0.2)  .. (proto.west);
\draw[secondary] (chembl.west).. controls +(-1.0,-0.5) and +( 1.0,0.5)  .. (onto.east);
\path (legend.north) (mesh.west) (drugchem.east);
\end{tikzpicture}
\end{adjustbox}
\caption{Unified resource$\rightarrow$modality map with typical fields. Central circles denote modalities, solid arrows are primary contributions and dashed arrows indicate contextual linkages (e.g., ontology crosswalks informing protocol text).}
\label{fig:resource_unified}
\end{figure*}

Fig.~\ref{fig:resource_unified} consolidates these sources into three modality channels used by our framework, such as \emph{Molecular (MOL)}, \emph{Protocol (PROTO)}, and \emph{Ontology (ONTO)}, together with a \emph{Labels} node used for supervised training/evaluation. Solid arrows denote \emph{primary} contributions: (i) ClinicalTrials.gov/AACT $\rightarrow$ Protocol (phase, enrollment, arms, comparators, endpoints, eligibility); (ii) MeSH and ICD-10 $\rightarrow$ Ontology (hierarchies and diagnostic semantics); (iii) DrugBank/PubChem and ChEMBL $\rightarrow$ Molecular (IDs, SMILES/InChI, targets, bioactivities); (iv) Gene Ontology/KEGG/Reactome $\rightarrow$ Ontology (mechanistic pathways); and (v) CTOD $\rightarrow$ Labels (NCT IDs, phase, outcome, therapeutic area descriptors, curated splits). Dashed arrows denote \emph{contextual} linkages that support textualization and cross-resource alignment, such as UMLS crosswalks feeding protocol term normalization and ChEMBL target/indication relationships reinforcing ontology grounding.

Methodologically, GNNs have been widely used to capture relationships among drugs, diseases, and targets in heterogeneous graphs for repurposing and outcome inference \cite{himmelstein2017systematic,wu2023biomedkg,wu2023kge,lee2020gnn,zhou2022gnn}. Recognizing that real-world trials interweave molecular, disease and protocol dimensions, \cite{fu2022hint} proposed \emph{HINT}, a hierarchical interaction network that models interactions at multiple levels (\emph{trial record}~$\leftrightarrow$~\emph{indication}~$\leftrightarrow$~\emph{molecule}), so that representation at one level is contextually conditioned by others. Ensemble strategies also sought robustness by combining heterogeneous learners in treatment domains \cite{sahoo2021survey}. Despite these advances, most frameworks still depend on modality-specific encoders, such as graph encoders for molecules, token encoders for text, bespoke projections for ontologies, thus limiting adaptability when onboarding new or underrepresented modalities and constraining end-to-end integration \cite{miotto2016deep,shickel2018deepehr}.

\begin{table*}[!t]
\centering
\caption{Public Biomedical Resources and What Each Contributes to Multimodal Trial Prediction}
\label{tab:resources}
\renewcommand{\arraystretch}{1.08}
\footnotesize
\begin{tabular}{p{0.17\textwidth} p{0.32\textwidth} p{0.30\textwidth} p{0.17\textwidth}}
\hline
\textbf{Resource} & \textbf{Core Content (What It Contributes)} & \textbf{Typical Fields Leveraged (Examples)} & \textbf{Modality Signals} \\
\hline
ClinicalTrials.gov / AACT \cite{clinicaltrials,ctti_aact,tasneem2012aact} & Structured protocol metadata and rich narratives; anchors outcome labels and design context across phases. & \emph{Phase, enrollment, arms, randomization/blinding, comparator (placebo/SoC), endpoints, eligibility, intervention descriptions}. & Tabular (metadata), Unstructured text (narratives) \\
\addlinespace[2pt]
DrugBank \cite{wishart2018drugbank} & Curated drug knowledge: identifiers, SMILES, substructures, pharmacologic class, MoA, targets. & \emph{SMILES/InChI, targets, mechanism of action, ATC class, synonyms}. & Graph/sequence (molecular), Structured text \\
\addlinespace[2pt]
PubChem \cite{kim2021pubchem,pubchem} & Large-scale chemical repository for structure normalization and computed properties. & \emph{Canonical SMILES, InChIKey, substructure fingerprints, physchem (e.g., logP)}. & Graph/sequence (molecular) \\
\addlinespace[1pt]
ChEMBL \cite{chembl} & Bioactivity database linking compounds to targets and indications; supports MoA grounding. & \emph{Targets, assays, bioactivities, indications}. & Structured molecular/target relations \\
\addlinespace[2pt]
MeSH / ICD-10 \cite{who_icd10} & Hierarchical disease ontologies for indication/comorbidity semantics. & \emph{Codes, preferred labels, tree positions, parent/child links, synonyms}. & Semi-structured ontology (hierarchy) \\
\addlinespace[2pt]
UMLS \cite{bodenreider2004unified} & Terminology integration and concept mappings across vocabularies; aids cross-resource linkage. & \emph{CUIs, synonymy, crosswalks (ICD/MeSH/etc.)}. & Ontology / concept graph \\
\addlinespace[2pt]
Gene Ontology / KEGG / Reactome \cite{ashburner2000gene,kanehisa2017kegg,jassal2020reactome} & Mechanistic context: functions, pathways, and processes linked to targets and diseases. & \emph{GO terms, pathway memberships, reactions}. & Pathway/ontology context \\
\addlinespace[2pt]
CTOD \cite{gao2024automatically} & Aggregated historical outcomes; benchmarking substrate for outcome prediction and cross-phase evaluation. & \emph{Trial IDs, phase, outcome (success/failure), therapeutic area, descriptors}. & Tabular (labels/metadata) \\
\hline
\end{tabular}
\end{table*}

\subsection{Challenges in Multimodal Integration}
Integrating clinical-trial data remains difficult due to four recurring challenges: (i) Representation mismatch: molecules are graphs of atoms and bonds, diseases are hierarchical ontologies, protocols are long free-text narratives, and trial designs are tabular; fusing such signals requires careful alignment across modality-specific encoders \cite{zitnik2018polypharmacy,shickel2018deepehr}; (ii) Encoder lock-in and onboarding: domain-specific encoders hinder rapid integration of new sources (e.g., imaging biomarkers or genomics), often requiring architectural changes and end-to-end retraining \cite{zhou2022gnn}; (iii) Phase-shift generalization: models tuned on Phase~II often underperform on Phase~III as cohort sizes, baseline risks, and endpoints change; fusion strategies that overweight ubiquitous narrative text can underuse molecular and ontology evidence and fail to carry across phases \cite{miotto2016deep}; and (iv) Balancing heterogeneous evidence: coverage is uneven (protocol text is nearly universal; high-quality molecular/ontology annotations are sparse), missing fields are common, and noise levels differ by modality, so the integrator must reweight inputs adaptively while remaining stable. Together, these challenges motivate mechanisms that can (a) map heterogeneous inputs into a \emph{common, auditable representation} to enable cross-modal reasoning, and (b) perform \emph{selective, robust fusion} that balances modalities rather than letting the most verbose source dominate.

\subsection{LLMs for Biomedical Integration and Trial Ops}
LLMs can standardize heterogeneous trial signals via schema-guided textualization and grounding, producing auditable narratives and structured slots that serve as inputs to downstream multimodal encoders and fusion models \cite{agarwal2025thinkjson,minaee2024llmsurvey,holley2024llms}. In practice, these narratives are embedded with modality-appropriate encoders (e.g., BERT-family biomedical models for clinical text) rather than treated as a final representation \cite{devlin2019bert,lee2020biobert,lin2025bridging}. This affords: \emph{auditability} (human-auditable narratives exposing functional groups, inclusion thresholds, comorbidity hierarchies), \emph{faithfulness} (schema-guided prompts and constrained templates that curb hallucination and enforce key fields), and \emph{adaptability} (new modalities require an encoder/adapter plus projection into the shared space) \cite{agarwal2025thinkjson}. A key enabler of faithful textualization is the use of \emph{standard vocabularies and pathway resources} to normalize entities and ground mechanistic context. UMLS aligns biomedical terminologies and supports cross-resource mapping \cite{bodenreider2004unified}; ICD-10 codifies diagnostic semantics for indications and comorbidities \cite{who_icd10}. Mechanistic knowledge comes from Gene Ontology, KEGG, and Reactome \cite{ashburner2000gene,kanehisa2017kegg,jassal2020reactome}. On the molecular side, SMILES provides a compact representation amenable to template-driven rendering \cite{weininger1988smiles}, and MoleculeNet curates tasks/benchmarks that inform transfer learning and feature validation \cite{wu2018moleculenet}. 

In our framework, these standards anchor field names and value spaces in the templates, improve entity linking across resources, and supply clinically meaningful descriptors that LLMs can leverage during controlled generation. LLMs are also impacting adjacent trial workflows: patient–trial matching \cite{jin2023matching}, automated screening support and RAG-enabled assistants in clinical operations \cite{unlu2024rag,wang2024gptclinical}, often implemented atop vector search and sentence embeddings \cite{reimers2019sentencebert,douze2024faiss,singh2024langchain}. Related multimodal outcome frameworks (e.g., \cite{zheng2025multimodal,zheng2025lifted}) highlight language as a universal layer but often assume coherent textual inputs already exist; our approach treats textualization itself as a primary engineering target and couples it with selective expert routing.

\subsection{Mixture-of-Experts for Scalable Specialization}

Mixture-of-Experts (MoE) offers a principled way to scale capacity while specializing to subspaces \cite{jacobs1991adaptive}. Essentially, sparsely-gated MoE layers route each input to a small subset of experts via learned gating, achieving high capacity at manageable compute \cite{shazeer2017outrageously,fedus2022switch}. Recent work explores softer routing/regularization to encourage balanced expert usage and stability \cite{puigcerver2023softmoe}, while comprehensive surveys review algorithmic choices and trade-offs in modern MoE systems, including applications in language models \cite{mu2025survey,cai2025moe}. This is appealing for clinical-trial outcome prediction because the salience of evidence shifts across indications and phases (e.g., molecular signals in oncology versus eligibility stringency in metabolic disease). For example, prior work such as LIFTED  applies a similar multimodal MoE architecture for trial outcome prediction \cite{zheng2025multimodal,zheng2025lifted}. Our proposed MMCTOP framework differs by (i) elevating schema-guided textualization (grounded in UMLS/ICD and pathway ontologies) as the unifying substrate instead of free-form or partially prompted narratives and (ii) conditioning sparse expert gating on drug and disease embeddings to promote clinically meaningful routing.

\section{Methodology} \label{sec:Methodology}

\subsection{Conceptual Framework}

\begin{figure*}[t]
  \centering
  \includegraphics[width=0.8\linewidth]{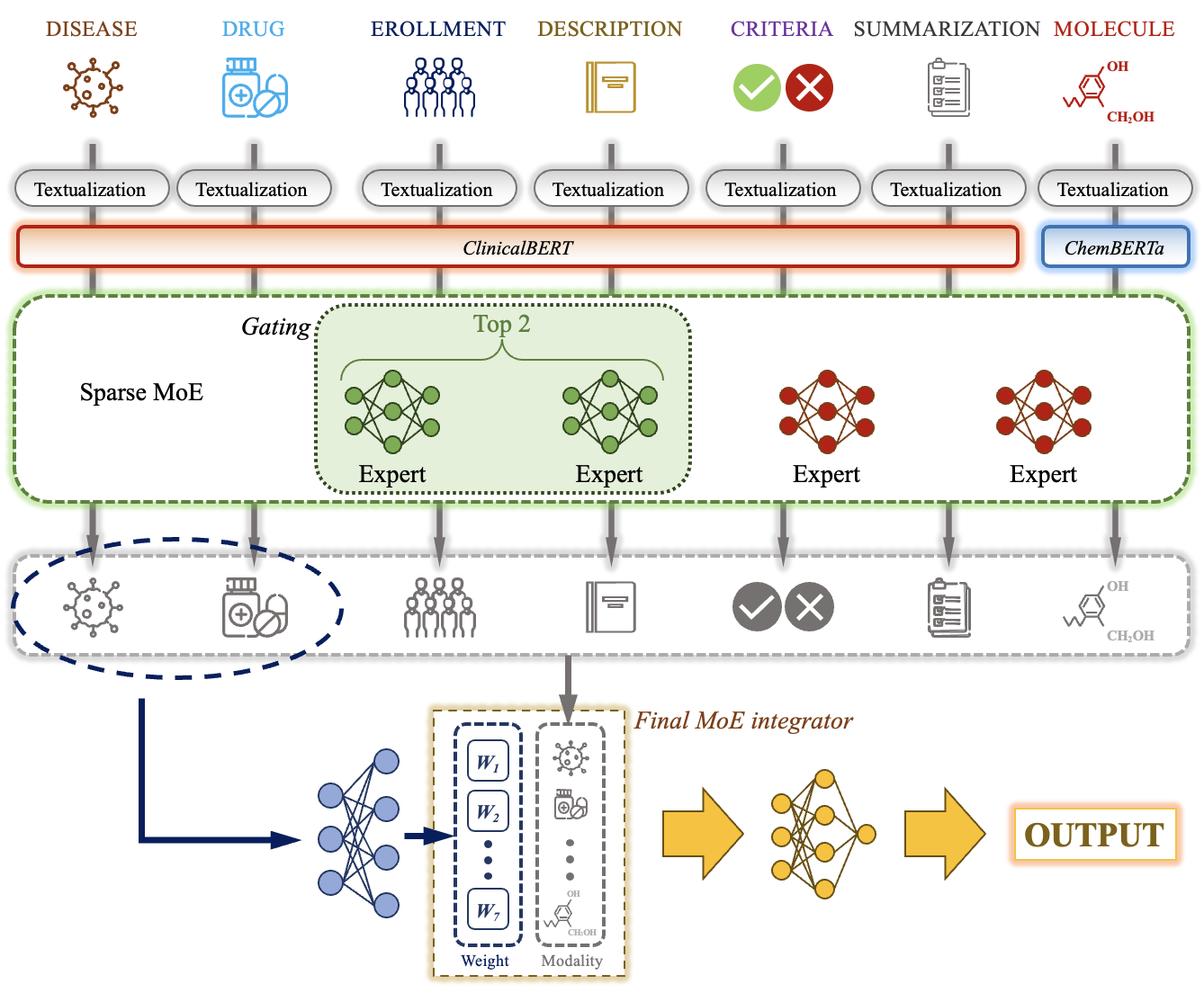}
  \caption{\textbf{MMCTOP architecture.}
Heterogeneous clinical-trial signals (disease, drug, enrollment, description, eligibility criteria, summaries, and molecular structure) are first standardized through \emph{schema-guided textualization and normalization}, producing auditable narrative artifacts and structured slots.
Clinical text modalities are encoded using a biomedical language model (ClinicalBERT), while molecular structures are encoded using a chemical language model (ChemBERTa).
Modality-specific embeddings are projected into a shared latent space and cached for computational efficiency.
A \emph{drug--disease--conditioned} sparse Mixture-of-Experts (SMoE) performs top-$k$ routing to specialized experts, followed by a \emph{final MoE integrator} that balances modality contributions and produces the trial outcome prediction.
}
  \label{fig:MMCTOP_pipeline}
\end{figure*}

Figure~\ref{fig:MMCTOP_pipeline} presents the MMCTOP pipeline for clinical trial outcome prediction. The framework resolves representation mismatch by \emph{standardizing heterogeneous inputs through schema-guided textualization} and then performing \emph{selective multimodal fusion} via a transformer-based SMoE. Beyond normalization, schema-guided textualization functions as an interpretable and auditable intermediate layer, producing human-readable artifacts that can be inspected, validated, and logged prior to representation learning, thereby mitigating black-box risks in multimodal biomedical fusion. New data modalities can be incorporated by introducing a corresponding encoder and mapping its outputs into the shared embedding space, without modifying the downstream fusion and decision layers.

First, heterogeneous trial signals, including molecular structures, protocol text and metadata, and disease ontology information are processed through schema-guided textualization and normalization. This stage produces auditable narrative artifacts and structured slots with explicit field ordering, controlled vocabularies, and deterministic decoding, together with cross-field consistency checks and ontology grounding. Accordingly, textualization is treated as a governance and interpretability layer that standardizes inputs and enforces schema conformance, while representation learning is delegated to modality-specific encoders. Second, standardized inputs are encoded using \emph{modality-specific encoders}. Clinical narratives (e.g., eligibility criteria, protocol descriptions, disease labels, and summaries) are embedded using a biomedical language model (ClinicalBERT), while molecular structures represented as SMILES are embedded using a chemical language model (ChemBERTa). SMILES strings are included verbatim during textualization to ensure faithful rendering and identifier consistency, but molecular signals are derived from ChemBERTa embeddings. All modality-specific embeddings are projected into a shared latent space and cached to improve computational efficiency and reproducibility. Third, the aligned embeddings are integrated using a sparse Mixture-of-Experts (SMoE) module. A drug--disease--conditioned gating network performs noisy top-$k$ routing, selecting a small subset of experts for each instance to enable scalable and context-dependent specialization across therapeutic areas and trial designs \cite{shazeer2017outrageously,fedus2022switch}. Fourth, a lightweight final MoE integrator aggregates expert outputs while enforcing modality-balance regularization to prevent dominance by any single evidence source, producing the trial outcome prediction.

\subsection{Data Sources and Modalities}
We use three modality families drawn from public resources listed in Table~\ref{tab:resources}. These sources provide complementary signals that are standardized and grounded \emph{during schema-guided textualization and preprocessing}, prior to modality-specific encoding and multimodal fusion.

\noindent\textbf{(i) Molecular.}
Small-molecule structures (SMILES/graphs) convey substructures, scaffolds, and pharmacologic context (mechanism of action and targets when available). We canonicalize identifiers (SMILES/InChI$\rightarrow$InChIKey), sanitize structures using RDKit, and map synonyms across DrugBank, PubChem, and ChEMBL.

\noindent\textbf{(ii) Protocol.}
Structured metadata (phase, enrollment, randomization/blinding, comparators/arms) and unstructured narratives (eligibility criteria, endpoints, intervention descriptions) are retrieved from ClinicalTrials.gov/AACT and processed as inputs to schema-guided textualization, including unit normalization, spelling harmonization, and sentence segmentation.

\noindent\textbf{(iii) Ontology.}
Indications and comorbidities are represented using MeSH and ICD preferred labels, hierarchical parent relations, and synonyms. UMLS Concept Unique Identifiers (CUIs) are used as crosswalks to harmonize disease semantics across resources and support consistent grounding during textualization.

%Each family is rendered into a faithful narrative and concatenated with segment separators into a single sequence for encoding.

\subsection{Natural-Language Templates and Textualization}
\label{subsec:textualization}
In MMCTOP, LLM-generated artifacts such as \textit{brief\_summary} and \textit{text\_description} are treated as standardized textual inputs to downstream biomedical language encoders rather than as a final unified representation. All clinical narratives produced by textualization are subsequently embedded using a domain-specific language model (ClinicalBERT). Molecular information is encoded exclusively via a chemical language model (ChemBERTa), with SMILES strings included in prompts to preserve chemical fidelity, contextual grounding, and identifier consistency, but not to replace molecular embedding. The primary function of textualization is therefore schema conformance, semantic grounding, and consistency enforcement prior to representation learning and multimodal fusion.

\begin{figure*}[t]
  \centering
  \includegraphics[width=0.8\linewidth]{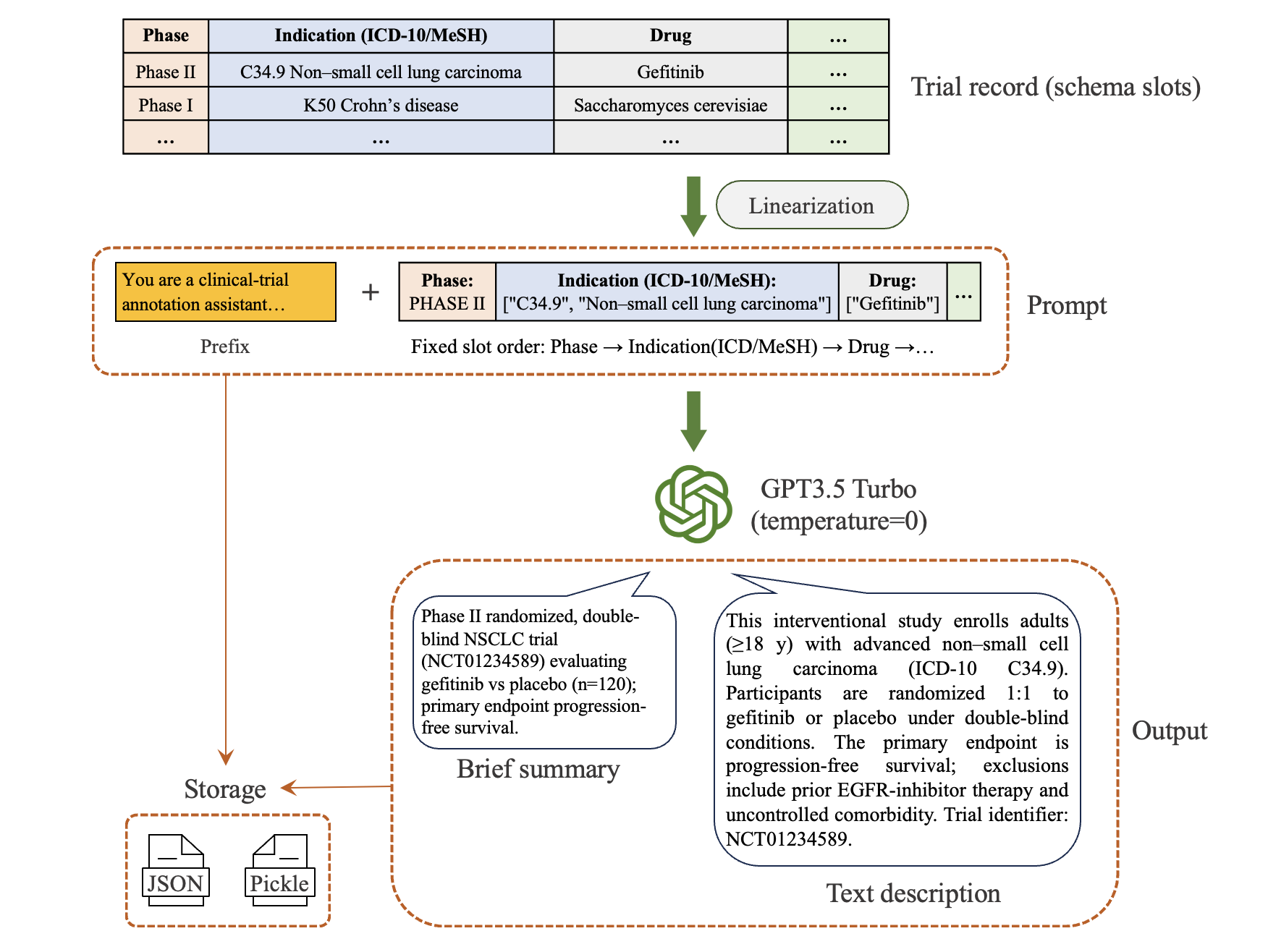}
  \caption{\textbf{Schema-guided textualization.} Structured clinical-trial records are \emph{linearized} into slot-ordered prompts (prefix + key:value pairs) that serve as deterministic inputs to the LLM (GPT-3.5-turbo; \(T{=}0\)). The model produces two artifacts: a \emph{brief summary} and a detailed \emph{text description}, both persisted as JSON (processed) and Pickle (raw) files for audit and reproducibility.}
  \label{fig:textualization}
\end{figure*}

MMCTOP performs a structured data–to–text transformation that converts heterogeneous clinical-trial records into descriptive natural-language narratives suitable for downstream encoding and multimodal fusion. As illustrated in Fig.~\ref{fig:textualization}, the process follows a schema-guided  textualization paradigm in which structured fields of each trial records are first linearized into slot-ordered prompts (schema prefix + \texttt{key:value} pairs + instruction suffix)~\cite{zheng2025multimodal}, which are subsequently passed to LLM (\emph{GPT-3.5-turbo}) for generation at zero temperature (\(T{=}0\)) to eliminate sampling variance.

\paragraph*{(1) Linearization.}
For each trial record \(i\) with structured fields indexed by \(k\), let \(c_{i,k}\) denote the slot (column name) and \(x_{i,k}\) its corresponding value. Each entry is mapped to a key–value pair:
\[
\mathrm{Linearize}(x_{i,k}) = \{\,c_{i,k} : x_{i,k}\,\}.
\]
The linearized prompt forms an ordered list spanning phase, disease, drug names, enrollment, arms/randomization/blinding, comparators, primary endpoints, indication codes/labels, comorbidities, and SMILES. This design enables the LLM to recognize schema structure explicitly, consistent with the formulation in~\cite{agarwal2025thinkjson}.

\paragraph*{(2) Prompt format and decoding.}
The prompt comprises three components, including a fixed schema prefix, the linearized slots, and a deterministic instruction suffix\footnote{Table~\ref{tab:prompt_examples} presents an example prompt, including the prefix specification and the two output modalities generated by GPT-3.5-turbo.}. Decoding is strictly deterministic (\(\text{temperature}=0\)), ensuring reproducible outputs. SMILES strings are inserted verbatim to preserve chemical fidelity, and long text fields (e.g., inclusion/exclusion criteria) are segmented into sentences. Two narrative outputs are generated:  
(i) a one-sentence \textit{brief\_summary} capturing the primary objective and design; and  
(ii) a fuller \textit{text\_description} describing population, design, comparator, and endpoints.

\paragraph*{(3) Grounding and quality control.}
Vocabulary is grounded to controlled resources to ensure domain fidelity: diseases are mapped to ICD/MeSH via UMLS CUIs, and molecular entities to DrugBank or ChEMBL (mechanism of action and targets). Meanwhile, we run several quality control checks, including (i) regex/JSON conformance checks (legal phases; integer enrollment; arm/comparator agreement; endpoint presence), (ii)  cross-field consistency (e.g., Phase~III $\Rightarrow$ minimum enrollment; ICD parent coherence with condition string), unit normalization, (iii) synonym canonicalization (e.g., \emph{NSCLC}$\leftrightarrow$\emph{non–small cell lung cancer}), and (iv) RDKit-based SMILES/InChI canonicalization. Single-slot inconsistencies are automatically repaired; irrecoverable records are rejected with reason codes.

\paragraph*{(4) Schema slots.}
Table~\ref{tab:template_slots} summarizes the schema structure used for MMCTOP textualization, grouping slots by modality and showing representative surface forms and primary data sources.

\begin{table*}[!t]
\centering
\caption{Schema-guided template slots and decoding constraints used in MMCTOP textualization.}
\label{tab:template_slots}
\renewcommand{\arraystretch}{1.05}
\footnotesize
\begin{tabular}{p{0.15\textwidth} p{0.25\textwidth} p{0.38\textwidth} p{0.18\textwidth}}
\hline
\textbf{Modality / Slot} & \textbf{Field Description} & \textbf{Example and Constraint} & \textbf{Primary Source} \\
\hline
\textbf{Molecular (MOL)} &  &  &  \\
\addlinespace[2pt]
\emph{scaffold, functional groups, substructures, ring systems, MoA, targets} &
Chemical structure and mechanism descriptors. &
\textit{Example:} ``Small molecule with benzene scaffold, two hydroxyls, tertiary amine; putative VEGFR inhibition.'' \newline
\textit{Constraint:} Canonicalized with RDKit; valid SMILES/InChI required. &
DrugBank, PubChem, ChEMBL \\
\addlinespace[6pt]

\textbf{Protocol (PROTO)} &  &  &  \\
\addlinespace[2pt]
\emph{phase} &
Clinical trial phase label. &
\textit{Example:} \texttt{phase: PHASE1, PHASE2} \newline
\textit{Constraint:} Must be one of \{I, II, III\}; coherent with enrollment size. &
ClinicalTrials.gov / AACT \\
\addlinespace[2pt]
\emph{arms / randomization / blinding / comparator / primary endpoint / enrollment / condition} &
Design metadata describing structure, control, and endpoints. &
\textit{Example:} ``Phase II randomized, double-blind, two arms (drug vs.\ placebo), 120 adults with NSCLC; primary endpoint PFS.'' \newline
\textit{Constraint:} Internal consistency required (arms$\leftrightarrow$comparator$\leftrightarrow$endpoint); enrollment must be a positive integer. &
ClinicalTrials.gov / AACT \\
\addlinespace[2pt]
\emph{criteria} &
Eligibility inclusion/exclusion narrative. &
\textit{Example:} ``Inclusion: 18–75 years, IBS confirmed by Rome IV; Exclusion: pregnancy, chronic alcoholism, eating disorder.'' \newline
\textit{Constraint:} Sentence-segmented; JSON-safe; deterministic decoding ($T{=}0$). &
ClinicalTrials.gov / AACT \\
\addlinespace[6pt]

\textbf{Ontology (ONTO)} &  &  &  \\
\addlinespace[2pt]
\emph{code $\rightarrow$ preferred label $\rightarrow$ parent; comorbidities / symptoms} &
Disease and comorbidity ontology fields. &
\textit{Example:} ``C0006826 (Non–Small Cell Lung Carcinoma), parent: Lung Neoplasms; comorbidities: COPD, hypertension.'' \newline
\textit{Constraint:} ICD/MeSH linkage via UMLS CUI; parent–child coherence enforced. &
MeSH / ICD / UMLS \\
\addlinespace[6pt]

\textbf{Generated outputs} &  &  &  \\
\addlinespace[2pt]
\emph{brief\_summary} &
One-sentence synopsis of the trial. &
\textit{Example:} ``The trial evaluates \textit{Saccharomyces cerevisiae} for Irritable Bowel Syndrome across phases I–II.'' \newline
\textit{Constraint:} $\leq$ 1 sentence; factual, no speculation. &
Generated (LLM) \\
\addlinespace[2pt]
\emph{text\_description} &
Full narrative description combining molecular, protocol, and ontology context. &
\textit{Example:} ``This is a clinical trial with phases 1 and 2 involving patients with IBS receiving \textit{Saccharomyces cerevisiae}...'' \newline
\textit{Constraint:} Deterministic decoding ($T{=}0$); factual; no paraphrasing beyond schema scope. &
Generated (LLM) \\
\hline
\end{tabular}
\end{table*}

\paragraph*{(5) Persistence and tokenization.}
All input--output pairs are logged, and both raw (pickle) and processed (JSON) artifacts are preserved for auditability.
For text modalities, we tokenize using the \emph{WordPiece} tokenizers associated with the underlying Transformer encoders (ClinicalBERT for clinical/protocol narratives and ChemBERTa's tokenizer for SMILES-derived sequences).
Per-segment caps are 128 (molecular), 512 (protocol), and 128 (ontology) tokens, with a global cap of 1{,}024 tokens.
Protocol truncation respects sentence boundaries and prioritizes endpoints and eligibility clauses.

\subsection{Predictive Modeling}
\label{subsec:predictive_modeling}

Following textualization, MMCTOP embeds each modality using domain-specific encoders (ClinicalBERT, ChemBERTa), projects embeddings into a shared space, and performs fusion and prediction using transformer+SMoE.

\subsubsection{Multimodal Embedding and Encoding Strategy}

At the core of MMCTOP, each input modality ranging from molecular descriptors to structured clinical metadata and natural-language narratives, is transformed into dense vector representations using Transformer-based, domain-specific encoders. This unified embedding approach preserves the semantic content of each modality and enables cross-domain reasoning across structured and unstructured data, including SMILES strings, drug and disease identifiers, eligibility criteria, enrollment statistics, and narrative protocol text.

\paragraph*{(1) Textual modality processing.}

Text-based components such as eligibility criteria, drug names, disease mentions, and descriptive summaries are tokenized using modality-specific tokenizers. The \textit{bert-base-cased} tokenizer is used for general biomedical text, while \textit{ClinicalBERT} (\texttt{medicalai/ClinicalBERT}) is employed for clinical narratives, including eligibility criteria and protocol sections~\cite{liu2025generalist}.

Eligibility criteria are segmented into inclusion and exclusion groups using rule-based phrase matching (\emph{``include,'' ``exclude,'' ``not eligible,''} etc.). Each sentence is encoded independently with \textit{ClinicalBERT}, which is pretrained on large-scale clinical corpora and configured for sequences up to 512 tokens. During encoding, a special \texttt{[CLS]} token is prepended and used as a semantic summary vector. The hidden state of the final \texttt{[CLS]} token from the last layer yields a 768-dimensional embedding per sentence.

To represent the full eligibility context, mean pooling is applied separately over the \texttt{[CLS]} embeddings from inclusion and exclusion sentences, producing two vectors
$e_{\mathrm{incl}}, e_{\mathrm{excl}} \in \mathbb{R}^{768}$.
These vectors are concatenated to form a single eligibility representation
$e_{\mathrm{elig}} = [e_{\mathrm{incl}} \| e_{\mathrm{excl}}] \in \mathbb{R}^{1536}$.
To ensure dimensional consistency with other modalities, the eligibility embedding is mapped into the shared latent space via a learned linear projection
$P_{\mathrm{elig}} \in \mathbb{R}^{768 \times 1536}$, yielding
$z_{\mathrm{elig}} = P_{\mathrm{elig}} e_{\mathrm{elig}} \in \mathbb{R}^{768}$.
Only the projected vector $z_{\mathrm{elig}}$ is used in downstream multimodal fusion.

This approach enhances contextual fidelity compared to earlier baselines such as \textit{Bio\_ClinicalBERT}~\cite{zheng2025multimodal}, while ensuring architectural consistency and preventing dimensional mismatch in the fusion layer.

\paragraph*{(2) Molecular representation.}
Molecular structures represented as SMILES strings are encoded using \textit{ChemBERTa}
(\texttt{seyonec/ChemBERTa-zinc-base-v1}), pretrained on the ZINC database for chemical
representation learning~\cite{chithrananda2020chemberta}. The hidden representation
associated with the \texttt{[CLS]} token is used as a fixed-length molecular embedding
$e_{\mathrm{mol}} \in \mathbb{R}^{768}$. Canonicalization using RDKit ensures structural
consistency across inputs. The molecular embedding is mapped into the shared multimodal
space via a learned linear projection, preserving pretrained chemical semantics while
enabling integration with other modalities in the downstream fusion stage.

\paragraph*{(3) Additional modality embeddings.}
All remaining textual and structured modalities, including drug names, disease labels,
protocol metadata, descriptive text, and enrollment information, are embedded using
\texttt{ClinicalBERT}. Each modality produces a 768-dimensional representation, which
is passed through a learned linear projection to ensure alignment within
the shared multimodal latent space.

\paragraph*{(4) Contextual encoding and modality structure.}
Contextual representations are produced internally by the pretrained encoders.
ClinicalBERT and ChemBERTa employ learned positional embeddings and self-attention to
capture intra-sequence dependencies within textual and molecular inputs, respectively.

For multimodal fusion, modality identity and structure are preserved at the embedding level through explicit modality grouping and projection into a shared latent space. When required by downstream fusion layers, modality indicators are used to distinguish molecular, protocol, and ontology representations, without re-encoding raw token sequences.

\vspace{0.5em}
\subsubsection{Multimodal Fusion Architecture}

To integrate modality-specific embeddings into a coherent predictive
representation, MMCTOP employs a sparse Mixture-of-Experts (SMoE) fusion
layer that dynamically allocates computation across specialized expert
subnetworks. Each expert is a feedforward neural module trained to
specialize in distinct cross-modality interaction patterns (e.g.,
molecular–disease, protocol–disease). Compared with dense concatenation
or cross-attention–based fusion, sparse MoE mitigates modality dominance
by verbose inputs (e.g., long protocol text) while preserving sparse but
high-impact biological signals from molecular embeddings, a critical
requirement for heterogeneous biomedical data.

A learned gating network computes expert selection scores via a linear
projection of the fused input representation followed by softmax
normalization. The gating function activates only the top-$k$ experts
(with $k=2$), minimizing inference cost while maintaining representational
diversity. The gating input is conditioned on drug and disease embeddings,
allowing expert selection to adapt to therapeutic context. Following
\cite{shazeer2017outrageously}, stochastic noise is injected during gating
to encourage balanced expert utilization and prevent mode collapse.

The outputs of the selected experts are aggregated through a
\emph{weighted summation}, with weights given by the gating probabilities.
This design enables context-dependent expert specialization while
preserving a fixed-dimensional fused representation.

To prevent expert under-utilization, an \emph{expert-importance loss}
penalizes imbalanced routing distributions, encouraging uniform
participation across experts and improving generalization.

\paragraph*{Auxiliary loss for expert utilization.}
To encourage balanced routing and avoid expert under-utilization, we apply the
standard Switch-style load-balancing loss~\cite{shazeer2017outrageously,fedus2022switch}.
Let $N$ denote the number of experts. For expert $i$, let $f_i$ be the fraction
of samples routed to expert $i$ in a minibatch, and let $P_i$ be the average
gating probability assigned to expert $i$. We define:
\begin{equation}
L_{\mathrm{imp}} = N \sum_{i=1}^{N} f_i \cdot P_i .
\end{equation}
The total training objective is
\begin{equation}
L = L_{\mathrm{bce}} + \lambda_{\mathrm{imp}} L_{\mathrm{imp}},
\end{equation}
where $\lambda_{\mathrm{imp}}$ controls the trade-off between predictive accuracy
and balanced expert utilization.

\paragraph*{Formal definition.}
Let $x_{\mathrm{mol}}$, $x_{\mathrm{proto}}$, and $x_{\mathrm{onto}}$
denote the molecular, protocol, and ontology inputs for a given trial
instance. Schema-guided textualization $T(\cdot)$ produces normalized,
auditable textual artifacts for each modality, which are subsequently
consumed by modality-specific encoders.

Each modality is encoded independently:
\[
\begin{aligned}
e_{\mathrm{mol}} &= E_{\mathrm{mol}}\!\left(x_{\mathrm{mol}}\right), \\
e_{\mathrm{proto}} &= E_{\mathrm{text}}\!\left(T_{\mathrm{proto}}(x_{\mathrm{proto}})\right), \\
e_{\mathrm{onto}} &= E_{\mathrm{text}}\!\left(T_{\mathrm{onto}}(x_{\mathrm{onto}})\right),
\end{aligned}
\]
where $E_{\mathrm{mol}}$ denotes the chemical language model (ChemBERTa)
and $E_{\mathrm{text}}$ denotes the biomedical language model
(ClinicalBERT).

Each embedding is projected into a shared latent space:
\[
z_m = P_m e_m, \quad m \in \{\mathrm{mol}, \mathrm{proto}, \mathrm{onto}\}.
\]

The aligned modality embeddings are concatenated to form the fusion input:
\[
h = [z_{\mathrm{mol}} \, \| \, z_{\mathrm{proto}} \, \| \, z_{\mathrm{onto}}].
\]

A sparse Mixture-of-Experts (MoE) layer with experts
$\{f_j\}_{j=1}^{E}$ and routing distribution $\pi(\cdot)$ computes:
\[
y = \sum_{j \in \mathcal{K}_k(h)} \pi_j(h)\, f_j(h),
\]
where $\mathcal{K}_k(h)$ denotes the top-$k$ experts selected by a
drug--disease--conditioned gating network.

Finally, the fused expert representation $y$ is concatenated with the
drug and disease embeddings and passed to a lightweight classifier:
\[
\ell = g\!\left([y \,\|\, z_{\mathrm{mol}} \,\|\, z_{\mathrm{onto}}]\right),
\qquad
\hat{p} = \sigma(\ell),
\]
where $g(\cdot)$ denotes a linear prediction head.

% --------- new subsection folding prior Experimental Setup ---------
\subsection{Experimental Setup} \label{sec:ExperimentalSetup}

\subsubsection{Datasets}
\paragraph*{TOP (Trial Outcome Prediction)~\cite{fu2022hint}}

The TOP dataset comprises 102{,}655 trial-phase records spanning Phases~I–III.\footnote{Phase~IV post-marketing studies were excluded as they lie beyond the developmental prediction scope~\cite{ratan2023applied}.}  
Each record corresponds to a specific trial--phase instance and includes 13 curated fields, covering trial title, phase, target diseases, molecular descriptors (SMILES), eligibility criteria, and a binary outcome label indicating trial success or failure. Because a single clinical trial may appear in multiple phases, the number of records exceeds the number of unique trials.

We retain only completed small-molecule interventional trials with valid molecular structures, disease codes, and phase annotations. Drug molecules are canonicalized to InChIKey format using RDKit, and disease labels are standardized via UMLS mappings across ICD and MeSH terminologies to ensure cross-dataset consistency.

Given the computational cost of schema-guided textualization and model training, we apply proportional stratified sampling based on the Clinical Classifications Software Refined (CCSR) taxonomy to preserve therapeutic-area representativeness. Each trial is assigned to its primary disease category using the first valid CCSR code, and sampling proportions are aligned with the full dataset’s therapeutic distribution. This procedure preserves outcome heterogeneity, such as systematically lower success rates in oncology relative to metabolic or infectious diseases, enabling unbiased evaluation~\cite{BIO2021}.

After filtering and stratified reduction, the final TOP subset used in our experiments contains 23{,}350 clinical trials, distributed across phases as follows: 7{,}736 Phase~I trials, 9{,}662 Phase~II trials, and 5{,}952 Phase~III trials. This design yields a statistically balanced representation of early- and mid-stage clinical development and supports robust generalization across therapeutic areas.

%\paragraph*{HINT.} Interventional trials spanning Phases~I–III with curated protocol fields and drug descriptors \cite{fu2022hinthierarchicalinteractionnetwork}. We retain completed trials with harmonized phases, standardize ICD/MeSH disease labels via UMLS, and canonicalize SMILES/InChI to InChIKey (RDKit). Where coverage is high, we use the full benchmark; when compute is constrained, we take stratified reductions preserving phase and therapeutic-area proportions (details in Appendix~S2).

\paragraph*{CTOD (Clinical Trial Outcome Database)~\cite{gao2024automatically}}

The CTOD dataset was derived from the AACT repository~\cite{ctti_aact}, integrating metadata, molecular structures, and outcome labels. The original collection contained 124{,}917 clinical trial records with 14 structured fields, including trial title, disease, drug description, SMILES representation, eligibility criteria, trial phase, enrollment and binary success/failure outcome labels.  

Consistent with the  procedures used to preprocess the TOP dataset, only completed small-molecule interventional trials reporting primary-outcome statistics and containing valid phase, drug, and molecular information were retained. Similarly, a proportional stratified sampling reduction was applied based on CCSR disease categories to maintain the therapeutic diversity and outcome balance of the dataset. The resulting CTOD dataset contained 16{,}225 trials (Phase~I: 5{,}022; Phase~II: 6{,}738; Phase~III: 4{,}465), all corresponding to early- and mid-stage small-molecule trials.

\subsubsection{Baselines}
We compare MMCTOP with several baseline families supported by the PyTrial framework~\cite{wang2023pytrial}. The first group includes traditional machine learning models such as Logistic Regression (LR), trained on structured trial features encompassing drug and protocol descriptors, and XGBoost (XGB), a gradient-boosted tree ensemble applied to tabular representations engineered without textual unification. A Multilayer Perceptron (MLP) serves as a shallow neural baseline trained on concatenated feature embeddings to capture limited nonlinear relationships. In addition, the HINT model serves as a reference multimodal benchmark predictor, integrating molecular and protocol representations through a hierarchical interaction network followed by a standard classification head. To assess the marginal contribution of each modality, unimodal or modality-specific variants are also evaluated individually, including models trained on only one input type, such as summarization, SMILES, eligibility criteria, description, enrollment, disease, or drug information.

\subsubsection{Model Evaluation}
Model performance is evaluated using  evaluation metrics including the Area Under the ROC Curve (AUC), Average Precision (PR), and F1-score, computed across all clinical trial phases and for each dataset (TOP and CTOD). These metrics assess both the discriminative capacity and precision of the models in identifying successful trial outcomes. For each dataset, the data are divided into independent training, validation, and testing sets, following an 80/10/10 stratified split by phase to ensure proportional representation across Phases~I–III. Performance is monitored exclusively on the validation set during training.
The model checkpoint with the highest validation AUC is selected for final evaluation, and the test set is used only once for reporting results. All experiments are repeated with fixed random seeds to ensure reproducibility. Computation is conducted on a single NVIDIA RTX~4080 GPU (16\,GB), and full model training, including ablation runs, follows identical hyperparameter and data split configurations.

\subsection{Ablation Studies}
\label{subsec:ablation_studies}
To evaluate the contribution of specific components within the MMCTOP architecture, two ablation variants are examined under identical data splits, optimization settings, and training configurations. The first variant, \emph{MMCTOP-AltGate}, modifies the gating mechanism in the SMoE layer by expanding its input from the original drug and disease embeddings to all available modality embeddings. This configuration tests whether incorporating broader contextual information influences expert routing dynamics and phase-transfer behavior. The second variant, \emph{MMCTOP-noNL}, removes the natural-language textualization component, excluding the language model–generated summarization and descriptive text from the multimodal input. This setup isolates the contribution of schema-guided textualization to the model’s representational capacity and predictive integration. Both variants are trained and evaluated using the same hyperparameters and stratified data partitions as the full model to ensure comparability.

\section{Results} \label{sec:Results}

\subsection{Performance Across Modalities}
\label{subsec:modality_performance}
\noindent
Table~\ref{tab:MMCTOP_modalities_merged} reports a modality-wise comparison between MMCTOP and single-modality variants on the \textsc{TOP} and \textsc{CTOD} datasets, illustrating the relative contribution of each input type (Summarization, SMILES, Description, Criteria, Enrollment, Diseases, and Drugs) across trial phases. Across both datasets and all phases, the full multimodal configuration consistently achieves the strongest overall performance in terms of PR, F1, and AUC, indicating that no single modality matches the predictive capacity obtained by jointly modeling heterogeneous biomedical evidence. Textual modalities such as Summarization, Description, and Criteria exhibit
moderate standalone predictive power, with PR values typically in the 70–83 range and AUC values around 58–62, reflecting their ability to capture high-level trial design and eligibility information. In contrast, structure-only
signals, including SMILES-based molecular representations and Enrollment statistics, perform substantially worse in isolation, particularly in early phases, with F1 scores often in the 20–30 range.

\begin{table*}[!t]
\centering
\caption{Modality-wise comparison against \textcc{MMCTOP (ours)} 
across datasets and trial phases. Metrics reported are PR, F1, and AUC. Best per column within each (Dataset, Phase) block is \textbf{bold}.}
\label{tab:MMCTOP_modalities_merged}
\scriptsize
\begin{adjustbox}{max width=\textwidth,center}
\begin{tabular}{llccc ccc ccc}
\toprule
 & & \multicolumn{3}{c}{\textbf{Phase I}} & \multicolumn{3}{c}{\textbf{Phase II}} & \multicolumn{3}{c}{\textbf{Phase III}} \\
\cmidrule(lr){3-5}\cmidrule(lr){6-8}\cmidrule(lr){9-11}
Dataset & Modality
& PR & F1 & AUC
& PR & F1 & AUC
& PR & F1 & AUC \\
\midrule
\multirow{8}{*}{\textsc{TOP}}
& Summarization & 82.47 & 66.83 & 61.92 & 65.90 & 53.26 & 57.88 & 73.64 & 56.12 & 54.90 \\
& SMILES        & 61.58 & 20.74 & 53.41 & 57.66 & 23.58 & 52.86 & 64.22 & 27.63 & 53.20 \\
& Description   & 70.96 & 44.31 & 58.27 & 66.98 & 51.73 & 57.95 & 72.81 & 55.42 & 55.40 \\
& Criteria      & 71.83 & 46.27 & 58.94 & 67.40 & 49.64 & 57.60 & 71.26 & 53.61 & 54.80 \\
& Enrollment    & 60.41 & 24.68 & 52.73 & 57.21 & 27.46 & 53.62 & 63.17 & 30.22 & 52.90 \\
& Diseases      & 74.62 & 54.18 & 60.86 & 66.35 & 48.57 & 58.20 & 69.92 & 52.11 & 55.10 \\
& Drugs         & 73.88 & 51.36 & 59.97 & 67.10 & 47.92 & 58.10 & 71.84 & 54.64 & 55.20 \\
& \textbf{MMCTOP (ours)}
& \textbf{86.88} & \textbf{87.50} & \textbf{69.14}
& \textbf{68.22} & \textbf{75.98} & \textbf{58.74}
& \textbf{77.75} & \textbf{80.27} & \textbf{56.28} \\
\midrule
\multirow{8}{*}{\textsc{CTOD}}
& Summarization & 59.37 & 42.18 & 55.42 & 74.83 & 58.91 & 61.26 & 80.71 & 63.44 & 65.53 \\
& SMILES        & 62.08 & 22.64 & 54.21 & 63.74 & 25.41 & 55.12 & 68.63 & 30.27 & 56.04 \\
& Description   & 72.61 & 50.74 & 60.33 & 78.42 & 57.28 & 63.22 & 82.36 & 61.18 & 66.07 \\
& Criteria      & 70.94 & 47.86 & 59.41 & 76.18 & 55.06 & 62.14 & 81.57 & 60.24 & 65.18 \\
& Enrollment    & 55.83 & 24.92 & 52.18 & 57.49 & 26.08 & 53.06 & 60.28 & 28.71 & 53.74 \\
& Diseases      & 69.86 & 48.63 & 58.37 & 77.12 & 56.11 & 62.03 & 80.38 & 61.07 & 65.09 \\
& Drugs         & 71.62 & 50.29 & 59.12 & 78.36 & 57.07 & 63.11 & 81.44 & 62.16 & 66.23 \\
& \textbf{MMCTOP (ours)}
& \textbf{91.58} & \textbf{86.01} & \textbf{77.52}
& \textbf{85.57} & \textbf{82.18} & \textbf{63.31}
& \textbf{91.15} & \textbf{90.30} & \textbf{72.91} \\
\bottomrule
\end{tabular}
\end{adjustbox}
\end{table*}

Importantly, the multimodal gains are not driven by any single dominant modality. Weak or sparse modalities, such as SMILES and Enrollment, show limited
discriminative capacity on their own, but their information becomes valuable when integrated with protocol narratives and disease context. This pattern suggests that multimodal fusion allows MMCTOP to recover complementary biological
and operational signals that are inaccessible to unimodal predictors.

On the \textsc{TOP} dataset, MMCTOP demonstrates consistent improvements over all single-modality baselines across Phases~I–III. Relative to the strongest unimodal models, AUC gains are typically on the order of 5–8 percentage points, with larger and more consistent improvements observed
in PR and F1. While absolute performance varies by phase, MMCTOP remains comparatively stable across the development lifecycle,
indicating that schema-guided textualization provides an effective interface for aligning molecular, clinical, and ontological information.

On the \textsc{CTOD} dataset, the advantages of multimodal fusion are even more pronounced, particularly in Phases~II and III where trial designs are more heterogeneous and outcome uncertainty is higher. In these settings, MMCTOP achieves substantial gains in both discrimination and F1 score compared with any single modality, highlighting the importance of jointly modeling molecular properties, eligibility constraints, and disease semantics for late-stage outcome prediction.

\subsection{Comparison with Baseline Models}
\label{subsec:baseline_performance}
\noindent
Table~\ref{tab:MMCTOP_baselines_merged} compares MMCTOP with traditional
machine learning and neural baselines on the \textsc{TOP} and \textsc{CTOD}
datasets. Across both datasets and trial phases, MMCTOP demonstrates the
strongest overall performance across most evaluation metrics, particularly in
terms of PR and AUC, indicating superior ranking and discrimination capability
under heterogeneous and noisy trial conditions.

On the \textsc{TOP} dataset, the proposed model consistently outperforms all
baselines in Phases~I and II, with the largest relative improvements observed in PR
and AUC. These early stages are characterized by limited sample sizes and higher
developmental uncertainty, and the observed gains indicate improved precision and
ranking reliability compared with conventional classifiers and prior multimodal
approaches. In Phase~III, while some baselines achieve competitive F1 scores,
MMCTOP maintains superior PR and AUC, suggesting more reliable
probability ranking despite class imbalance.

On the \textsc{CTOD} dataset, MMCTOP exhibits the widest performance
margins, achieving the strongest results among all evaluated models across nearly
all metrics and phases. Notably, in Phase~III, the model attains substantially
higher PR, F1, and AUC than all baselines, indicating that the combination of
multimodal encoding and sparse expert routing effectively captures complex
dependencies among drug properties, eligibility constraints, and trial design
features in late-stage development.

Overall, these results demonstrate that MMCTOP integrates heterogeneous
biomedical evidence more effectively than traditional machine learning models,
neural baselines, and prior multimodal methods, while generalizing robustly across
datasets and trial phases.

\begin{table*}[!t]
\centering
\caption{Baselines vs.\ \textcc{MMCTOP (ours)} across datasets and phases. Metrics are PR, F1, and AUC. Best per column within each (Dataset, Phase) block is \textbf{bold}.}
\label{tab:MMCTOP_baselines_merged}
\scriptsize
\begin{adjustbox}{max width=\textwidth,center}
\begin{tabular}{llccc ccc ccc}
\toprule
 &  & \multicolumn{3}{c}{\textbf{Phase I}} & \multicolumn{3}{c}{\textbf{Phase II}} & \multicolumn{3}{c}{\textbf{Phase III}} \\
\cmidrule(lr){3-5}\cmidrule(lr){6-8}\cmidrule(lr){9-11}
\textbf{Dataset} & \textbf{Model} & PR & F1 & AUC & PR & F1 & AUC & PR & F1 & AUC \\
\midrule
\multirow{5}{*}{\textsc{TOP}}
& LR              & 78.10 & 87.40 & 51.70 & 60.30 & 74.50 & 48.50 & 72.40 & \textbf{83.90} & 50.50 \\
& MLP             & 77.60 & 87.20 & 51.80 & 62.10 & 74.50 & 52.30 & 71.70 & 83.50 & 49.60 \\
& XGB             & 77.40 & 87.20 & 48.40 & 58.70 & 74.30 & 52.70 & 72.10 & 83.50 & 51.30 \\
& HINT & 79.70 & 87.10 & 54.70 & 61.20 & 74.40 & 52.70 & 74.80 & 83.40 & 56.10 \\
& \textbf{MMCTOP (ours)} & \textbf{86.88} & \textbf{87.50} & \textbf{69.14} & \textbf{68.22} & \textbf{75.98} & \textbf{58.74} & \textbf{77.75} & 80.27 & \textbf{56.28} \\
\midrule
\multirow{5}{*}{\textsc{CTOD}}
& LR              & 85.60 & 83.90 & 70.10 & 80.80 & 80.70 & 61.00 & 84.10 & 85.20 & 69.30 \\
& MLP             & 86.00 & 85.50 & 70.30 & 78.40 & 81.90 & 61.80 & 85.60 & 88.30 & 71.50 \\
& XGB             & 85.80 & 84.20 & 74.50 & 80.20 & \textbf{82.60} & 61.90 & 85.10 & 88.90 & 72.40 \\
& HINT & 83.10 & 84.70 & 66.40 & 77.20 & 80.50 & 58.40 & 83.00 & 85.40 & 67.50 \\
& \textbf{MMCTOP (ours)} & \textbf{91.58} & \textbf{86.01} & \textbf{77.52} & \textbf{85.57} & 82.18 & \textbf{63.31} & \textbf{91.15} & \textbf{90.30} & \textbf{72.91} \\
\bottomrule
\end{tabular}
\end{adjustbox}
\end{table*}

Table~\ref{tab:MMCTOP_ablation_merged} reports ablation results across datasets and
trial phases, evaluating the contributions of schema-guided textualization and
drug--disease--conditioned expert routing. Removing schema-guided textualization
(\textcc{MMCTOP-noNL}) yields a mixed effect that depends on dataset, phase, and
metric: on \textsc{TOP}, \textcc{MMCTOP-noNL} underperforms \textcc{MMCTOP (ours)}
across phases on PR/F1/AUC, indicating that schema-conformant narrative artifacts
improve multimodal integration in this benchmark; on \textsc{CTOD}, however,
\textcc{MMCTOP-noNL} remains competitive and achieves the strongest Phase~III AUC,
suggesting that the benefit of textualization is not uniform across datasets and
may interact with dataset-specific label structure and protocol-field coverage.
Expanding the gating input to include all modalities (\textcc{MMCTOP-AltGate})
produces performance that is also metric-dependent: it can match or exceed the full
model on Phase~III F1 in both datasets, while \textcc{MMCTOP (ours)} retains stronger
PR/AUC in several phase blocks. Overall, the ablations indicate that both (i)
schema-guided textualization and (ii) constrained, context-conditioned routing can
affect performance, but their gains are not monotonic; instead, they trade off
precision-ranking (PR/AUC) and thresholded balance (F1) differently across phases.

\begin{table*}[!t]
\centering
\caption{Ablations across datasets and phases. \textcc{MMCTOP-AltGate} uses all modalities as gating inputs; \textcc{MMCTOP-noNL} removes LLM textualization.}
\label{tab:MMCTOP_ablation_merged}
\scriptsize
\begin{adjustbox}{max width=\textwidth,center}
\begin{tabular}{llccc ccc ccc}
\toprule
 &  & \multicolumn{3}{c}{\textbf{Phase I}} & \multicolumn{3}{c}{\textbf{Phase II}} & \multicolumn{3}{c}{\textbf{Phase III}} \\
\cmidrule(lr){3-5}\cmidrule(lr){6-8}\cmidrule(lr){9-11}
\textbf{Dataset} & \textbf{Variant} & PR & F1 & AUC & PR & F1 & AUC & PR & F1 & AUC \\
\midrule
\multirow{3}{*}{\textsc{TOP}}
& MMCTOP-AltGate & 83.45 & 87.11 & 64.12 & 66.90 & 74.37 & 55.54 & 74.13 & \textbf{81.78} & 51.70 \\
& MMCTOP-noNL    & 82.04 & 86.12 & 63.39 & 64.18 & 73.92 & 53.97 & 74.49 & 80.97 & 53.89 \\
& \textbf{MMCTOP (ours)} & \textbf{86.88} & \textbf{87.50} & \textbf{69.14} & \textbf{68.22} & \textbf{75.98} & \textbf{58.74} & \textbf{77.75} & 80.27 & \textbf{56.28} \\
\midrule
\multirow{3}{*}{\textsc{CTOD}}
& MMCTOP-AltGate & 88.31 & 81.64 & 74.93 & 81.11 & 77.23 & 60.58 & 89.05 & \textbf{91.03} & 72.59 \\
& MMCTOP-noNL    & 87.25 & 79.88 & 73.66 & 79.26 & 75.91 & 59.03 & \textbf{91.50} & 90.54 & \textbf{76.48} \\
& \textbf{MMCTOP (ours)} & \textbf{91.58} & \textbf{86.01} & \textbf{77.52} & \textbf{85.57} & \textbf{82.18} & \textbf{63.31} & 91.15 & 90.30 & 72.91 \\
\bottomrule
\end{tabular}
\end{adjustbox}
\end{table*}

% If you exported reliability diagrams, include them here; otherwise omit calibration figures.

\section{Discussion} \label{sec:Discussion}

\subsection{Scientific and Practical Implications}
The results demonstrate that schema-guided textualization, combined with domain-specific encoders and selective expert routing, improves clinical trial outcome prediction across heterogeneous biomedical modalities, including molecular structure, protocol narratives, and disease ontologies. MMCTOP consistently outperforms unimodal and non-textual multimodal baselines on both the \textsc{TOP} and \textsc{CTOD} datasets, with the largest gains observed in late-phase trials where design complexity and contextual interactions are most pronounced.

Ablation analyses further indicate that textualization functions as a critical normalization and grounding mechanism: removing the LLM-based narrative layer significantly reduces predictive performance, while selective drug–disease gating stabilizes fusion and prevents over-reliance on any single modality. These findings suggest that the primary contribution of natural-language artifacts lies in enabling coherent multimodal integration and conditional specialization under explicit architectural control, rather than in providing post-hoc explanatory signals.

From a practical perspective, the model’s discriminative properties have concrete implications for R\&D decision-making. Given that Phase~III trials typically cost between \$11.5 and \$52.9 million per study~\cite{sertkaya2016costs}, avoiding even a small fraction of failed trials could yield substantial cost savings.

Beyond aggregate performance gains, MMCTOP reveals domain-specific patterns consistent with established clinical knowledge. In oncology trials, molecular–textual interactions enable the model to link drug–target complexity and stringent eligibility criteria with elevated failure likelihood, aligning with known challenges in dose-finding and biomarker stratification. In cardiovascular studies, MMCTOP detects enrollment-to-endpoint inconsistencies often observed in large, multicenter trials. These patterns demonstrate that the model not only predicts outcomes but also captures biologically and operationally meaningful relationships across therapeutic areas, underscoring its potential to guide data-driven decision-making in disease-focused R\&D portfolios.

For translational integration, the model’s probabilistic outputs can be incorporated into existing clinical development workflows. For example, model-predicted risk rankings can inform early portfolio triage, adaptive sample-size recalibration, and go/no-go review processes. Although this study focuses on retrospective validation, the next step toward regulatory plausibility is prospective testing within simulated portfolio environments or adaptive-design feasibility assessments, ensuring that interpretability aligns with decision accountability in real-world trial governance.

Although this study relies on public clinical trial data, the proposed architecture is compatible with privacy-sensitive deployment settings. In particular, the textualization layer can act as a sanitization and governance boundary, enabling removal or redaction of protected health information prior to embedding and downstream modeling. This separation between raw clinical text and learned representations supports compliance-aware deployment in real-world clinical and regulatory environments.

\subsection{Cross-Phase Generalization and Broader Impact}
\noindent
A notable observation from cross-phase experiments is that MMCTOP trained on early-phase data generalizes well to predicting late-phase (Phase~III) outcomes. This behavior suggests that the model learns transferable semantic patterns linking molecular mechanisms and eligibility narratives to ultimate clinical success.
Such cross-phase generalization provides a foundation for forward-looking risk assessment, supporting adaptive portfolio governance, early reprioritization of compounds, and strategic de-risking of clinical pipelines.
Beyond pharmaceutical R\&D, the methodology exemplifies how narrative-based multimodal unification can advance biomedical AI systems more broadly, providing a framework applicable to other high-stakes prediction tasks such as trial site selection, patient recruitment forecasting, or drug repurposing evaluation.

\iffalse
\subsection{Limitations}
Several limitations temper these conclusions and point to areas for improvement. Data availability also constrains ceiling performance. In CTOD, key ADMET attributes are absent, even though such features are known to correlate with success; the model thus relies more heavily on proxy signals in molecular and protocol narratives. Label and feature imbalance is substantial, with success cases outnumbering failures in several indications; we mitigated this with inverse-frequency class weights, stratified splits, and threshold tuning on validation, but rare-event sensitivity remains challenging. 

The textualization step introduces its own risks. Although schema-first prompts, constrained decoding, and vocabulary whitelists reduced drift, we still observed occasional hallucination-like artifacts during template generation, particularly in long eligibility passages. These were addressed by strict regex/JSON conformance checks, cross-field consistency rules, and rejection or repair of offending records; nevertheless, any automated narrative extraction must be treated as a source of uncertainty. Finally, portions of the template generation relied on proprietary LLMs, which raises cost and reproducibility concerns for some settings. While the downstream encoder and MoE components are fully reproducible, future iterations should prefer high-quality open biomedical LLMs to lower cost and improve transparency.
\fi

\subsection{Future Directions}
\noindent
Several extensions follow naturally. Integrating real-world evidence (e.g., electronic health records, post-marketing safety data) could complement registry-based information and improve generalization to practical settings. Expanding from trial-level to patient-level prediction would enable individualized success and safety estimation, provided privacy-preserving mechanisms are employed.
From a modeling perspective, transitioning to open-source LLMs fine-tuned on biomedical corpora can reduce dependency on proprietary systems, while adapter-based tuning could enable disease- or modality-specific specialization at low computational cost.
Future work should also incorporate structured ADMET (Absorption, Distribution, Metabolism, Excretion, and Toxicity) and pharmacokinetic features where available, and conduct simulation-based validations of decision thresholds under realistic portfolio management scenarios.
These directions collectively aim to extend MMCTOP from a predictive research framework to an operational tool for data-driven clinical development governance.

Finally, as multimodal integration expands, privacy and data governance remain essential. Future versions of MMCTOP can incorporate federated or privacy-preserving learning to enable multi-institutional collaboration without centralizing patient-level data. De-identification pipelines and audit logs for schema transformations could further strengthen data security and regulatory compliance, aligning with the ethical standards emphasized in biomedical AI deployment.

\subsection{Limitations}
Despite strong empirical performance, MMCTOP has several limitations.
First, the schema-guided textualization step assumes a minimum level of
structured input quality; trials with severely incomplete or noisy records
may still suffer from residual textualization errors, although regex-based
validators and cross-field consistency checks mitigate this risk.
Second, while sparse expert routing improves interpretability at the routing
level, the internal decision logic of individual experts remains opaque.
Finally, the use of proprietary LLMs for textualization introduces potential
cost and reproducibility constraints, motivating future work on distilling
these capabilities into open-source biomedical language models.

\section{Conclusion} \label{sec:Conclusion}

\noindent
This study presents MMCTOP, a multimodal clinical trial outcome prediction framework that integrates heterogeneous biomedical evidence, including molecular descriptors, clinical protocol narratives, and disease ontologies, through schema-guided textualization, domain-specific encoding, and selective expert-based fusion. The model employs LLM-based textualization as an upstream normalization and governance layer, followed by modality-appropriate transformers and a sparse Mixture-of-Experts (SMoE) architecture for conditional specialization and robust prediction. Across both the \textsc{TOP} and \textsc{CTOD} datasets, MMCTOP consistently outperforms unimodal and traditional multimodal baselines while achieving improved cross-phase generalization and robust discrimination across trial phases. Ablation analyses confirm that the two central design choices: (i) schema-guided textualization for standardized, auditable input construction and (ii) selective drug–disease–conditioned expert routing, are critical for model stability and discriminative performance, particularly in late-phase trials where protocol complexity is highest.

\section*{Data and Code Availability}
We used public datasets from ClinicalTrials.gov (AACT), TOP, CTOD, DrugBank, PubChem, and MeSH/ICD.
Template schemas, validators, data splits, and training scripts will be released upon acceptance to ensure full reproducibility.

\section*{Ethics Statement}
This study analyzes publicly available, non-identifiable trial records and ontologies. No human subjects research or protected health information was used.

\bibliographystyle{IEEEtran}
%\section*{References}
\bibliography{biblio}

%\newpage
\appendix
\begin{table*}[t]
\centering
\caption{Exact prompt components used in MMCTOP textualization (prefix, linearized slots, and instructions suffix).}
\label{tab:prompt_examples}
\renewcommand{\arraystretch}{1.05}
\footnotesize
\begin{tabular}{|p{0.22\textwidth}|p{0.73\textwidth}|}
\hline
\textbf{Component} & \textbf{Content} \\
\hline
\textbf{Schema prefix (fixed)} &
\begin{minipage}[t]{\linewidth}\ttfamily\raggedright
You are a clinical-trial annotation assistant. You are given normalized clinical-trial fields as key:value slots in the following fixed order: \\
phase; diseases (ICD-10/MeSH); drugs; smiles; icdcode; criteria (Inclusion/Exclusion). \\
Use the values as facts; do not invent or infer missing data. \\
Use preferred ICD/MeSH labels and include UMLS CUI when available in parentheses. \\
Keep units and numerics unchanged; normalize spelling; no citations in the output.
\end{minipage}
\\
\hline
\textbf{Linearized slots (example)} &
\begin{minipage}[t]{\linewidth}\ttfamily\raggedright
phase: PHASE1, PHASE2; \\
diseases: ['Irritable Bowel Syndrome']; \\
drugs: ['saccharomyces cerevisiae']; \\
smiles: [\detokenize{CC(C)CCC[C@@H](C)[C@@]1([H])CC[C@@]\2([H])C(CCC[C@]12C)=CC=C1C[C@@H]\(O)CCC1=C}]; \\
icdcode: ['D6861']; \\
criteria: Inclusion Criteria: * Male and female patients between 18 and 75 years of age, * Patients having confirmed IBS according to Rome IV criteria (newly and previously non-responder to treatment), * Pain/discomfort score $>$1 and $<$6 on 0--7 scale in the 7 days preceding inclusion, * Not hypersensitive to any ingredient. \\
Exclusion Criteria: * Organic intestinal disease (Crohn’s, ulcerative colitis), * Pregnancy, * Treatments likely to influence IBS (antidepressants, opioids, narcotic analgesics), * Chronic alcoholism, vegetarian or vegan regimens, * Eating disorders (anorexia/bulimia), * Documented food allergies.
\end{minipage}
\\
\hline
\textbf{Instructions suffix (fixed)} &
\begin{minipage}[t]{\linewidth}\ttfamily\raggedright
Style: factual, concise, no speculation. Keep slot order semantics in your wording. \\
If a value is missing or out of vocabulary, write \textquotesingle unknown\textquotesingle. \\
Emit two artifacts: \\
(A) \textbf{brief\_summary}: exactly one sentence summarizing phase(s), indication, intervention, comparator (if present), and primary endpoint. \\
(B) \textbf{text\_description}: a short paragraph (3--5 sentences) covering population, design (randomization/blinding/arms), comparator, primary endpoint, and key eligibility themes. \\
Do not copy eligibility criteria verbatim; summarize key inclusion and exclusion
themes in a deterministic manner while preserving all numeric thresholds,
units, and logical constraints (e.g., inequalities, temporal windows).
Use ICD/MeSH preferred labels and UMLS CUI when provided. \\
Decoding: temperature=0; no sampling.
\end{minipage}
\\
\hline
\end{tabular}
\end{table*}

\end{document}